\documentclass[letterpaper, 10 pt, conference]{ieeeconf}  

\IEEEoverridecommandlockouts                              

\usepackage{cite}
\usepackage{amsmath,amssymb,amsfonts}
\usepackage{bm}
\usepackage{algorithmic}
\usepackage{graphicx}
\usepackage{textcomp}
\usepackage{epstopdf}
\usepackage{CJKutf8}

\title{\LARGE \bf
UEVAVD: A Dataset for Developing UAV's Eye View \\ Active Object Detection
}

\author{Xinhua Jiang$^{1}$, Tianpeng Liu$^{1*}$, Li Liu$^{1}$, Zhen Liu$^{1}$, and Yongxiang Liu$^{1}$
\thanks{$^{1}$All authors are with College of Electronic Science, National University of Defense Technology, Changsha, 421007, China.}
\thanks{*∗Corresponding author: Tianpeng Liu (email: everliutianpeng@sina.cn)}%
}

\begin{document}
\begin{CJK}{UTF8}{gbsn}

\maketitle
\thispagestyle{empty}
\pagestyle{empty}

\begin{abstract}

Occlusion is a longstanding difficulty that challenges the UAV-based object detection. Many works address this problem by adapting the detection model. However, few of them exploit that the UAV could fundamentally improve detection performance by changing its viewpoint. Active Object Detection (AOD) offers an effective way to achieve this purpose. Through Deep Reinforcement Learning (DRL), AOD endows the UAV with the ability of autonomous path planning to search for the observation that is more conducive to target identification. Unfortunately, there exists no available dataset for developing the UAV AOD method. To fill this gap, we released a UAV's eye view active vision dataset named UEVAVD and hope it can facilitate research on the UAV AOD problem. Additionally, we improve the existing DRL-based AOD method by incorporating the inductive bias when learning the state representation. First, due to the partial observability, we use the gated recurrent unit to extract state representations from the observation sequence instead of the single-view observation. Second, we pre-decompose the scene with the Segment Anything Model (SAM) and filter out the irrelevant information with the derived masks. With these practices, the agent could learn an active viewing policy with better generalization capability. The effectiveness of our innovations is validated by the experiments on the UEVAVD dataset. Our dataset will soon be available at https://github.com/Leo000ooo/UEVAVD$\_$dataset.

\end{abstract}

\section{INTRODUCTION}

\label{sec:introduction}
In recent years, Unmanned Aerial Vehicles (UAVs) have made a big splash in many applications like traffic monitoring \cite{01}, industrial facilities inspection \cite{02}, and post-disaster search and rescue \cite{03} owing to its high flexibility and strong maneuverability. Behind these applications, target detection is an indispensable key technology \cite{1}, which aims to locate and identify targets from aerial images, providing important prior information for subsequent actions. Since the renaissance of deep learning, the Deep Neural Network (DNN) based methods, such as Faster-RCNN \cite{2}, YOLO \cite{3}, SSD \cite{4}, and their variants have gradually become the mainstream in the field of UAV target detection \cite{5}. However, in the air-to-ground scenario, detecting a target would inevitably encounter challenge of the occlusion among terrestrial objects, which greatly diminishes the detection performance \cite{6}. To address this difficulty, most existing methods make adaptive improvements to the detection model. For example, literature \cite{7} uses the Soft-NMS method to suppress redundant prediction frames during post-processing to alleviate the occlusion problem. However, the detector's ability of anti-occlusion is still far from satisfactory \cite{8}. Fundamentally, the quality of the input data limits the upper bound of detection performance. If the UAV could autonomously change the viewing angle to get the observation more conducive to identifying the target, the detection performance would greatly improve. To achieve this, Active Object Detection (AOD) is a practical way.

\begin{figure}[!t]
	\centering
	\includegraphics[width=3.2in]{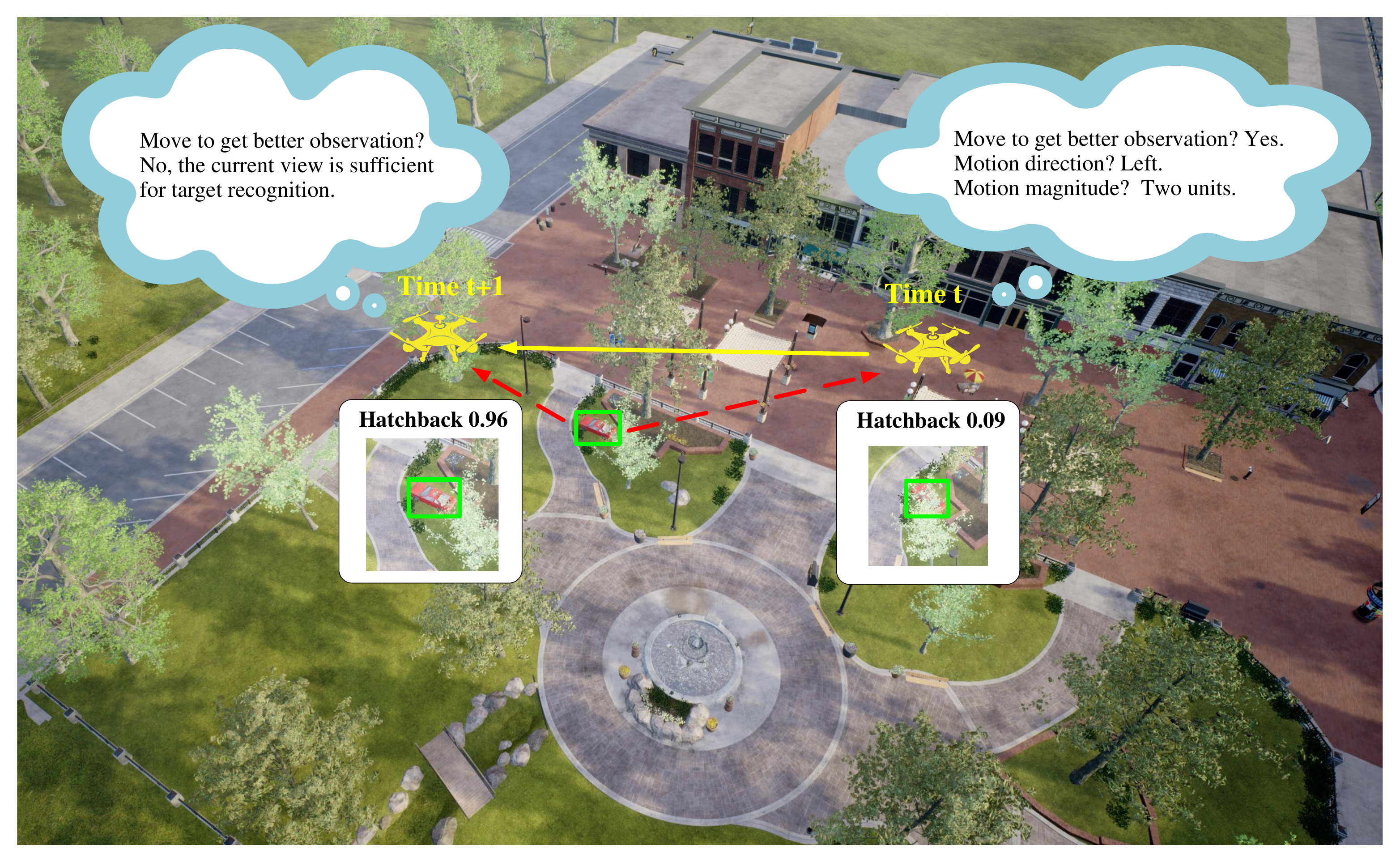}
	\caption{A demonstration of the AOD decision process by the UAV platform. The UAV's policy network gives moving instructions based on observations. It can autonomously decide whether to move and how to move to acquire an ideal observation while minimizing the movement cost.}
	\label{fig1}
\end{figure}

AOD is a subfield that utilizes active vision in target detection, and its objective is to improve the detection result by endowing the mobile sensing platform the ability to actively adjust the viewing angle to better identify the target whose location is given at first \cite{11}. As shown in Fig. \ref{fig1}, the UAV cannot determine the target identity under some viewpoints due to occlusion. At this time, the UAV has to decide whether to move and how to move to acquire an ideal observation while minimizing the movement cost. 

Deep Reinforcement Learning (DRL) has now become a mainstream framework for solving the AOD problems \cite{12}. Ammirato et al. first used the REINFORCE algorithm for AOD tasks and released the Active Vision Dataset (AVD) collected in indoor environments for developing and benchmarking active vision algorithms \cite{15}. Han et al. solved the AOD problem by a Dualing Deep Q-learning Network (DualingDQN) with prioritized experience replay for the first time \cite{16}. Fang et al. used self-supervised representation learning to improve the sample efficiency of the DRL method \cite{11}. Based on literature \cite{17}, Liu et al. incorporated the target crop into the state representation and designed a novel reward function to help the robot approach the target object more smoothly \cite{18}.

However, existing AOD researches mainly focu on indoor robotic applications, neglecting the outdoor air-to-ground circumstances. For the indoor environment, there are several datasets concerning active vision published, such as the AVD \cite{15}, T-LESS \cite{19}, and R3ED \cite{20} dataset, all of which collect the multi-view target images indoor under various occlusion conditions. However, there is no similar counterpart in the scenario of the UAV's eye view AOD. Although the existing datasets (e.g., VisDrone-DET dataset \cite{21}) for UAV-based target detection commonly cover abundant environment settings and targets, they cannot be used for studying the air-to-ground AOD problem because of lacking densely collected multi-view images over the targets.

To fill this gap, we release a UAV's Eye View Active Vision Dataset (UEVAVD) and hope it can facilitate research on the AOD problem in the air-to-ground scenario. The dataset is collected in the simulated environment constructed by Unreal Engine (UE), focusing on five types of vehicular targets. UAV observes each target under different environment settings (different occlusions, different terrains), and its predefined sampling positions are densely and regularly arranged over the target. Various combinations of the samples can be seen as the continuous observation result by the UAV while flying along different routes, laying a foundation for future research on the UAV AOD problem. 

Besides, given that a good state representation can dramatically strengthen the effectiveness, robustness, and generalization capability of the agent’s policy \cite{22}, we improve the existing AOD method by incorporating inductive biases into the state representation learning process. Specifically, for the AOD task, two aspects of prior knowledge can be exploited. First, due to the partial observability, the state representation of the agent needs to be extracted from the historical observation sequence instead of relying on the single-view observation. Therefore, we utilize a combination of the convolutional neural network and the Gated Recurrent Unit (GRU) to extract state representation from the observation sequence. Second, what should be stressed are the characteristics of the target itself in terms of its appearance, pose, and the positional relationship between the target and its surroundings, whereas features such as the color and texture of the terrestrial objects can be discarded. Therefore, we pre-decompose the scene using the powerful Segment Anything Model (SAM) and filter out the irrelevant information with the masks derived. With these practices, the backbone of the decision network could learn better state representation, thus improving the performance of the agent’s policy.

The main contributions of this paper are summarized as follows.
\begin{enumerate}
	\item{We release a new dataset, UEVAVD, aiming to promote the research on the UAV's eye view AOD problem. On this basis, we could find out how to better exploit the UAV's autonomy and maneuverability to overcome difficulties like occlusion in UAV-based object detection.}
	\item{We improve the existing DRL-based AOD method by incorporating the inductive bias while learning the state representation. With the practice of scene pre-decomposition and  memory based state estimation, the policy learned by the agent could get stronger generalization capability and perform better in the testing environment.}
\end{enumerate}

\section{DATA COLLECTION}

We first construct the whole simulation environment with UE and use the AirSim plugin to get observations from the UAV platform. This section provides detailed descriptions of the targets, environments, sampling process, and the statistical overview of the UEVAVD dataset.

\subsection{Targets and Environment Settings}
\begin{figure}[!ht]
	\centering
	\includegraphics[width=3.2in]{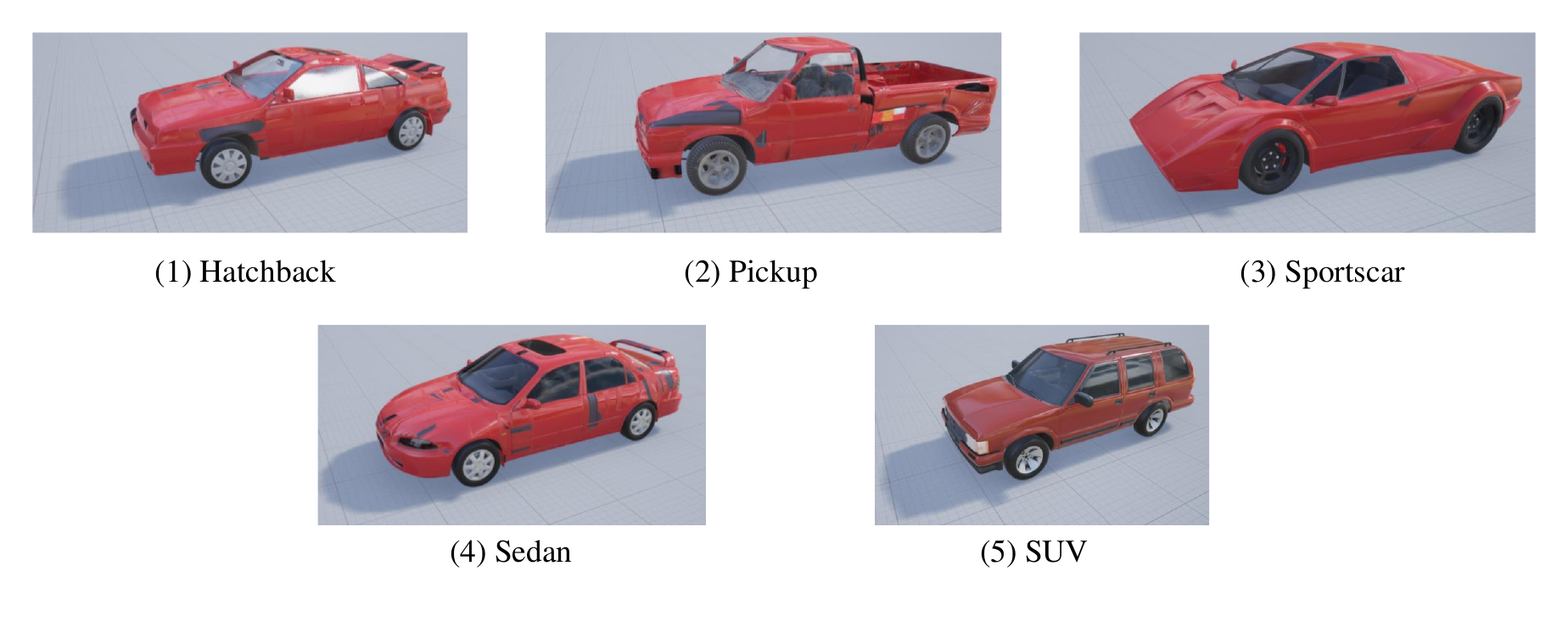}
	\caption{The five vehicle targets used in collecting the UEVAVD dataset.}
	\label{fig2}
\end{figure}
Our dataset focuses on vehicle targets in urban and woodland terrains. As shown in Fig. \ref{fig2}, we choose five types of vehicle targets from online resources and integrate them into our project. To avoid the classifier easily distinguishing targets solely with color information, we standardize the colors and textures of Hatchback, Pickup, Sedan, and SUV vehicles to match that of the Sportscar. The environment used for sample collection is as demonstrated in Fig. \ref{fig1}.
\begin{figure}[!t]
	\centering
	\includegraphics[width=3.2in]{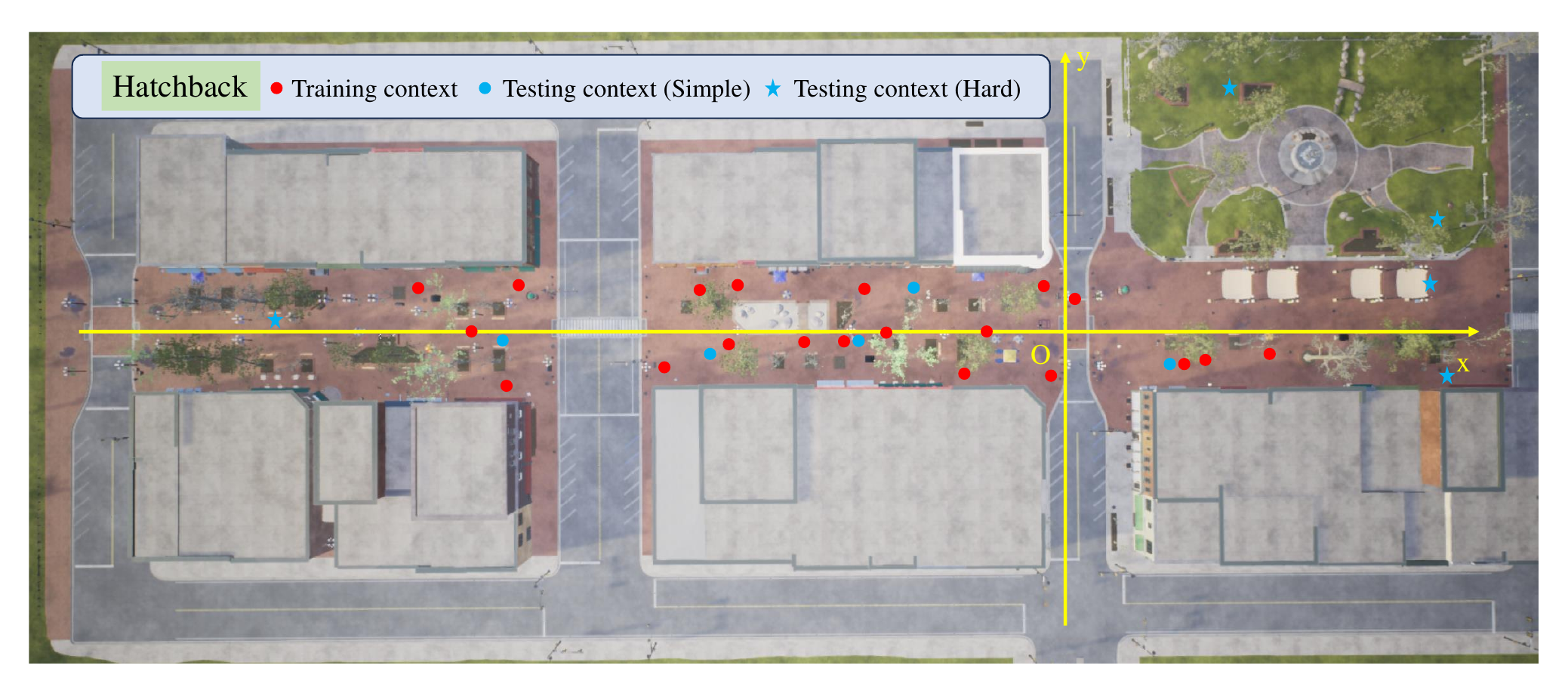}
	\caption{The sample distribution of target's location under three kinds of contexts. }
	\label{fig3}
\end{figure}

To ensure the richness and variety of our dataset,  we randomly distributed the targets' locations around the whole scene. Fig. \ref{fig3} shows the overhead view of the scene, where the target locations are denoted with points or stars, and each location corresponds to an environment setting or context. Each kind of target is placed at 30 different locations. Namely, there are 30 contexts for each target type where the background, illumination, and target occlusion vary. In Fig. \ref{fig3}, the red points correspond to the training contexts where we collected the training set, and the testing contexts are denoted with the blue points or stars, whose difference lies in that the latter more obviously deviate from the training contexts in terms of target position distribution. Besides, the ground objects around the target are supposed to be complex and varying enough. In this regard, we generally place the target next to buildings or trees. For simplicity, we make all the targets' orientations identical during data collection.

\subsection{Sampling Process}
\begin{figure}[!t]
	\centering
	\includegraphics[width=3.2in]{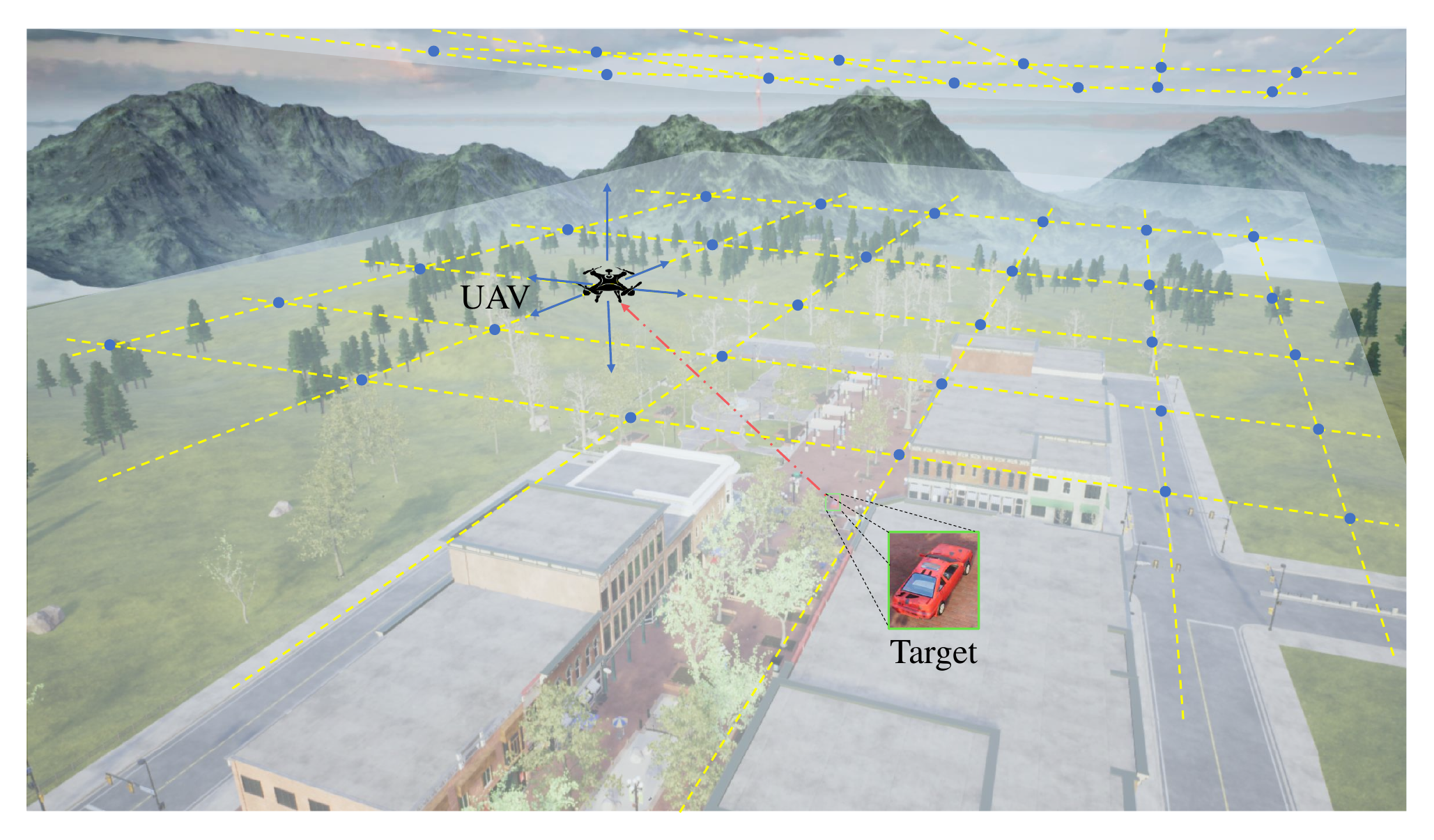}
	\caption{A demonstration of the predefined sampling points distribution and the UAV's moving directions. The lower airspace over the target is divided into discrete sections, and we record observations of the area of interest when the UAV is at these sampling points. }
	\label{fig4}
\end{figure}

In each scenario, the UAV observes the target at specific sampling points evenly distributed over the area of interest. Figure 4 shows the UAV's sampling points and the designated motion directions. Assuming the target location is the origin, let us consider a sampling point with coordinates $({x_s},{y_s},{z_s})$ in meters, and its distance from the origin is expressed as:
\begin{equation}
	\label{eq1}
	\left\{ \begin{array}{l}
		- 50 \le {x_s} \le 50,{\kern 1pt} {\kern 1pt} {\kern 1pt} {\kern 1pt} {x_s} \in \mathbb{Z}\\
		- 50 \le {y_s} \le 50,{\kern 1pt} {\kern 1pt} {\kern 1pt} {\kern 1pt} {y_s} \in \mathbb{Z}\\
		60 \le {z_s} \le 100,{\kern 1pt} {\kern 1pt} {\kern 1pt} {\kern 1pt} {z_s} \in \mathbb{Z}
	\end{array} \right..
\end{equation}
The minimum distance between the neighboring sampling points is 10 m. We assume that once the target is located, the UAV could automatically adjust the camera to track the target and make it show at the center of the aerial image.

With the AirSim plugin, we can obtain the original RGB image and the groundtruth segmentation image of the whole scene under the UAV's perspective, whose size is $4096 \times 2160$ pixels. Since the target only occupies a very small part of the whole original image and what the AOD task needs is the description of the target and its surroundings, we crop the original RGB images to obtain $256 \times 256$-large slices centered on the target. We also derive the groundtruth bounding boxes using the original RGB and segmentation images. The bounding box is denoted by ${\left[ {{x_{tl}},{y_{tl}},{x_{lr}},{y_{lr}}} \right]^T}$, where $({x_{tl}},{y_{tl}})$ and $({x_{lr}},{y_{lr}})$ are the coordinates of the top left and lower right point, respectively. If the target is completely blocked by obstacles, its bounding box is given by ${\left[ {0,0,0,0} \right]^T}$.
\subsection{Dataset Overview}
\begin{figure}[!t]
	\centering
	\includegraphics[width=3.4in]{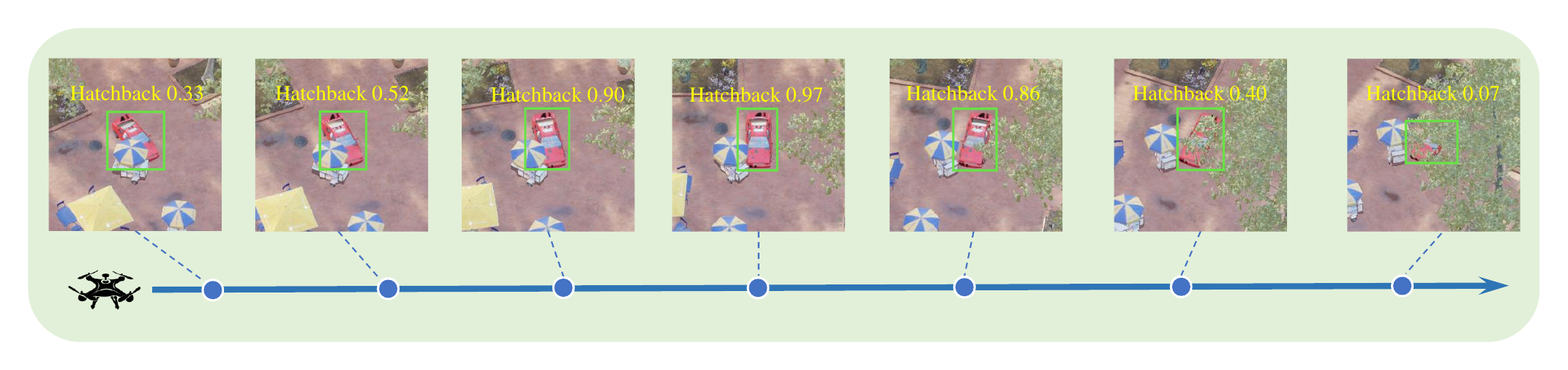}
	\caption{A sample sequence illustrating how the observation and recognition result change with the view variation. }
	\label{fig6}
\end{figure}

\begin{table}[!t]
	\caption{Details of the Proposed UEVAVD Dataset}
	\label{tab1}
	\renewcommand\arraystretch{1.2}
	\setlength{\tabcolsep}{0.1mm}{
		\begin{tabular}{|cccc|}
			\hline
			\multicolumn{4}{|c|}{UEVAVD Dataset}                                                                                                   \\ \hline
			\multicolumn{1}{|c|}{Split}             & \multicolumn{1}{c|}{Training Set} & \multicolumn{1}{c|}{Test Set (Simple)} & Test Set (Hard) \\ \hline
			\multicolumn{1}{|c|}{Target}            & \multicolumn{3}{c|}{Hatchback, Pickup, Sports Car, Sedan, SUV}                               \\ \hline
			\multicolumn{1}{|c|}{Terrain}           & \multicolumn{2}{c|}{Urban}                               & \multicolumn{1}{c|}{Urban, Woodland}                          \\ \hline
			\multicolumn{1}{|c|}{Context ID} &
			\multicolumn{1}{c|}{\begin{tabular}[c]{@{}c@{}}01, 02, ..., 20,\\ 31, 32, ..., 50,\\ 61, 62, ..., 80,\\ 91, 92, ..., 110,\\ 120, 121, ..., 140\end{tabular}} &
			\multicolumn{1}{c|}{\begin{tabular}[c]{@{}c@{}}21, 22, ..., 25,\\ 51, 52, ..., 55,\\ 81, 82, ..., 85,\\ 111, 112, ..., 115,\\ 141, 142, ..., 145\end{tabular}} &
			\begin{tabular}[c]{@{}c@{}}26, 27, ..., 30,\\ 56, 57, ..., 80, \\ 86, 87, ..., 90, \\ 116, 117, ..., 120, \\ 146, 147, ..., 150\end{tabular} \\ \hline
			\multicolumn{1}{|c|}{Context Number}    & \multicolumn{1}{c|}{100}          & \multicolumn{1}{c|}{25}                & 25              \\ \hline
			\multicolumn{1}{|c|}{Image Number}      & \multicolumn{1}{c|}{60500}        & \multicolumn{1}{c|}{15125}             & 15125           \\ \hline
			\multicolumn{1}{|c|}{Annotation Number} & \multicolumn{1}{c|}{60500}        & \multicolumn{1}{c|}{15125}             & 15125           \\ \hline
		\end{tabular}
	}
\end{table}
Table. \ref{tab1} provides the statistical overview of the UEVAVD dataset. It contains the multi-view imaging results of five types of vehicle targets under different environmental settings. The UEVAVD dataset contains multi-view imaging results with different extents of occlusion, their combinations can be used for simulating the continuous target observations under the moving UAV's view. Fig. \ref{fig6} tells that with the varying observation angle, the classification result changes simultaneously. In some cases, the occlusion is too severe to correctly classify the target, while sometimes, a single observation is sufficient for target recognition. Therefore, the key problem in AOD is finding the `good' view with the shortest movement path based on the current state to increase recognition accuracy efficiently.

\section{METHODOLOGY}

\subsection{Problem Formulation}
\label{problem_Formulation}
We model the AOD problem as a Partially Observable Markovian Decision Process (POMDP), which is usually represented by a seven-tuple $\left\langle {{\cal S},{\cal A},{\cal O},T,\Omega ,R,\gamma } \right\rangle $. ${\cal S}$ is the state set of the agent, and $\bm{s} \in {\cal S}$ corresponds to the state representation extracted from the UAV observation. ${\cal A}$ denotes the action set, and $\bm{a} \in {\cal A}$ is the predefined action, consisting of action type $a_{type}$ and action range $a_{range}$, where ${a_{type}} \in \{ forward,backward,left,right,down,up,stop\}$, and ${a_{range}} \in \{ 1,2,...,A\} $. ${\cal O}$ denotes the observation set. $T({\bm{s}}'|{\bm{s}},{\bm{a}})$ is the state transition function. $\Omega (\bm{o}|\bm{s},\bm{a})$ is the observation function, denoting the probability that the agent receiving observation $\bm{o}$ after taking action $\bm{a}$ at state $\bm{s}$. Observation $\bm{o}$ includes the aerial image ${\bm{I}}$ and the bounding box $\bm{b}$. In AOD task, the target's initial position is given beforehand, so its bounding box in subsequent frames can be obtained trivially through the tracking algorithm \cite{17}. $\gamma$ is the discount factor while calculating the return. $R({\bm{s}},{\bm{a}})$ is the reward function, whose design should take into consideration the detection accuracy, decision steps, and the movement path of UAV, so in this paper, the reward function is defined by:
\begin{equation}
	\label{eq2}
	R({\bm{s}},{\bm{a}}) = \left\{ \begin{array}{l}
		{\xi _1} - \sigma  \cdot {a_{range}},{\kern 1pt} {\kern 1pt} {\kern 1pt} {\kern 1pt} {\kern 1pt} if{\kern 1pt} {\kern 1pt} {\kern 1pt} {\kern 1pt} flag = 1,{\kern 1pt} {\kern 1pt} {\kern 1pt} {\kern 1pt} {\kern 1pt} reco = 1\\
		- {\xi _1} - \sigma  \cdot {a_{range}},{\kern 1pt} {\kern 1pt} {\kern 1pt} {\kern 1pt} {\kern 1pt} if{\kern 1pt} {\kern 1pt} {\kern 1pt} {\kern 1pt} flag = 1,{\kern 1pt} {\kern 1pt} {\kern 1pt} {\kern 1pt} {\kern 1pt} reco = 0\\
		- {\xi _2} - \sigma  \cdot {a_{range}},{\kern 1pt} {\kern 1pt} {\kern 1pt} {\kern 1pt} if{\kern 1pt} {\kern 1pt} {\kern 1pt} {\kern 1pt} flag = 0
	\end{array} \right..
\end{equation}
When ${a_{type}} = stop$ or $t = {T_{\max }} - 1$, the variable $flag$ is set to 1, denoting the end of an episode, or the agent needs to go on making decisions. If the target is correctly identified, the variable $reco$ is set to 1, or $reco=0$. ${\xi _1}$ and ${\xi _2}$ are both positive reward constants, and $\sigma$ is the coefficient used to control the weight of action range, which is defined by:
\begin{equation}
	\label{eq3}
	\sigma  = \left\{ \begin{array}{l}
		0,{\kern 1pt} {\kern 1pt} {\kern 1pt} {\kern 1pt} if{\kern 1pt} {\kern 1pt} {\kern 1pt} {a_{type}} = stop{\kern 1pt} \\
		C,{\kern 1pt} {\kern 1pt} {\kern 1pt} {\kern 1pt} else, {\kern 1pt} {\kern 1pt} {\kern 1pt} {\kern 1pt} C \geq 0 
	\end{array} \right..
\end{equation}

On this basis, the agent's optimal observation policy $\pi^*$ can be obtained by solving the following optimization problem:
\begin{equation}
	\label{eq4}
	\pi ^* = \mathop {\arg \max }\limits_{\pi  \in \Pi } {\mathbb{E}_{{\bm{s}} \sim \rho ({\bm{s}})}}\left[ {r({\bm{s}})} \right],
\end{equation}
where $\pi ({{\bm{a}}_t}|{{\bm{s}}_t})$ represents the agent's policy, and $\Pi$ denotes the policy set. $\rho ({\bm{s}})$ is the distribution of the initial state. $r({\bm{s}})$ is the expected return within an episode (we assume there are ${T_{\max }}$ decision steps in one episode), which is calculated by:
\begin{equation}
	\label{eq5}
	\resizebox{.9\hsize}{!}
	{$r({\bm{s}}) := {\mathbb{E}_{{{\bm{a}}_t} \!\sim\! \pi ({{\bm{a}}_t}|{{\bm{s}}_t}),{s_{t + 1}} \!\sim\! T({{\bm{s}}_{t + 1}}|{{\bm{s}}_t},{{\bm{a}}_t})}}\left[ {\!\sum\limits_{t = 0}^{{T_{\max }} - 1}\! {{\gamma ^t}R({{\bm{s}}_t},{{\bm{a}}_t})} {\left| {{{\bm{s}}_0} \!=\! {\bm{s}}} \right.}} \right]$}.
\end{equation} 

Since the agent's decision model needs to be trained and tested in multiple contexts, to quantify the Zero-Shot Generalization (ZSG) capability of the agent's policy on the UEVAVD dataset, we introduce the Contextual Markovian Decision Process (CMDP) \cite{23}, a special POMDP with the concept of context, expressed by $\left\langle {{\cal S}',{\cal A},{\cal C},{\cal O},T,\Omega ,R,p,\gamma } \right\rangle$. ${\cal C}$ is the context set, and the context variable $\bm{c} \in {\cal C}$ specifies the target's location. $p(\bm{c})$ denotes its probability distribution. Besides, ${\cal S}'$ means the latent state set, and there is a mapping that ${\cal S}' \times {\cal C} \to {\cal S}$. Within each episode, the context variable would remain unchanged, namely, if ${{\bm{c}}_1} \ne {{\bm{c}}_2}$, then $T\left( {({\bm{s}}{'_2},{{\bm{c}}_2})|({\bm{s}}{'_1},{{\bm{c}}_1}),a} \right) = 0$.

For the CMDP ${\cal M}$ with the context set ${\cal C}$, the expected return of the policy $\pi$ is defined by:
\begin{equation}
	\label{eq6}
	r(\pi ,{\cal M}|{\cal C}): = {\mathbb{E}_{{\bm{c}} \sim p({\bm{c}})}}\left[ {r(\pi ,{{\cal M}_{\bm{c}}})} \right],
\end{equation}
where ${\cal M}_{\bm{c}}$ is the CMDP when the context is fixed to ${\bm{c}}$. Let ${{\cal C}_{train}}$, ${{\cal C}_{test}}$ be the training context set and the test context respectively. By learning from numerous interaction experiences, the agent could obtain the optimal policy $\pi _{train}^*$ on the training environments until the return value comes to convergence, namely:
\begin{equation}
	\label{eq7}
	\pi _{train}^* = \mathop {\arg \max }\limits_{\pi  \in \Pi } r(\pi ,{\cal M}|{{\cal C}_{train}}).
\end{equation} 
We adopt the metric of $r(\pi ,{\cal M}|{{\cal C}_{test}})$ and $GenGap(\pi )$ \cite{28} to evaluate ZSG capability. For $\pi _{train}^*$, $GenGap(\pi _{train}^*)$ is calculated by:
\begin{equation}
	\label{eq8}
	\resizebox{.9\hsize}{!}
	{$GenGap(\pi _{train}^*) = r(\pi _{train}^*,{\cal M}|{{\cal C}_{train}}) - r(\pi _{train}^*,{\cal M}|{{\cal C}_{test}})$},
\end{equation}
the smaller $GenGap(\pi )$ and the higher $r(\pi ,{\cal M}|{{\cal C}_{test}})$ imply that the policy holds a better generalization capability. 

\subsection{Inductive Biases Enhanced AOD Method}
\label{overview}
\begin{figure*}[!t]
	\centering
	\includegraphics[width=6in]{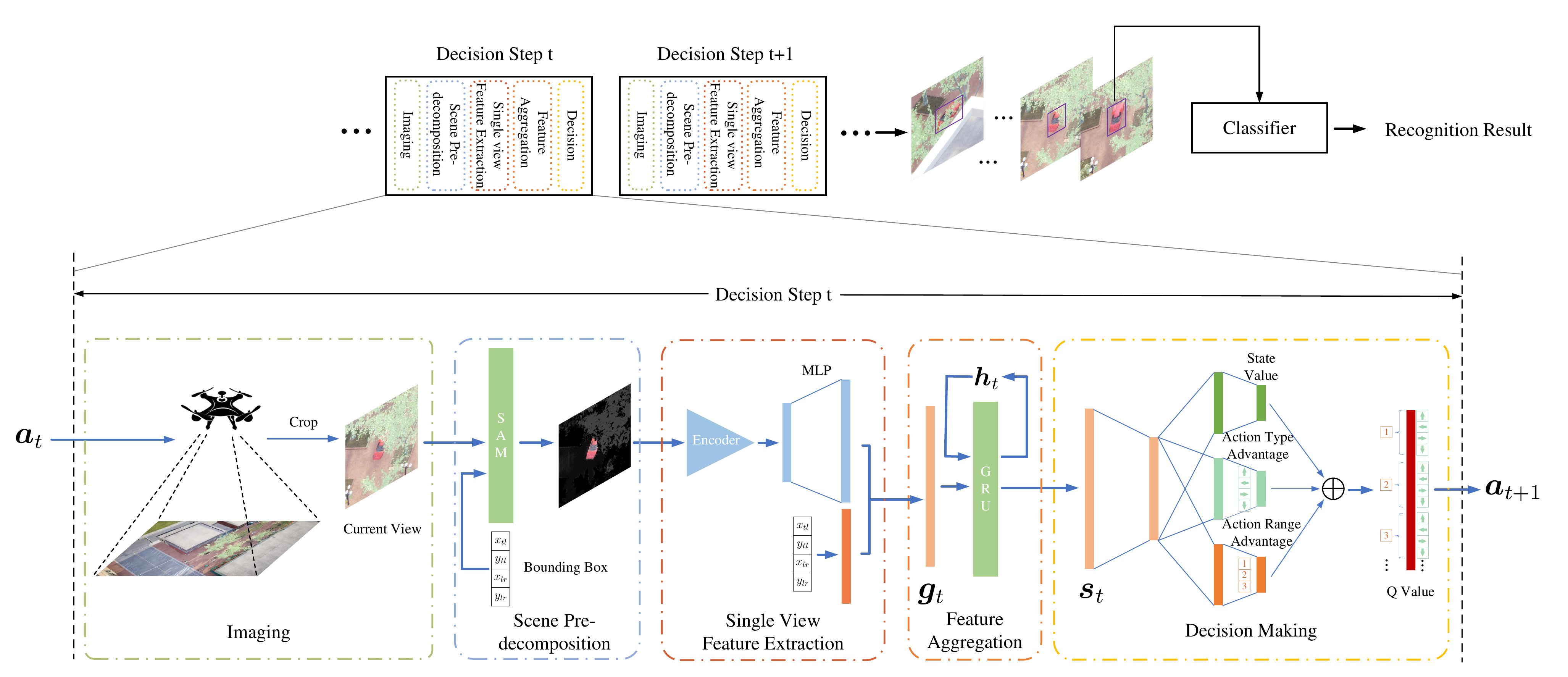}
	\caption{The inference flowchart of the proposed IBE-MAP method. The top row depicts the high-level inference workflow. The bottom part describes the details within a decision step. The operations of scene pre-decomposition and feature aggregation correspond to the added inductive biases, which help the backbone of the policy network to learn a better state representation, thus supporting the agent to obtain a better active viewing policy.}
	\label{fig7}
\end{figure*}
Compared to the former similar works \cite{11,12,15, 16,17,18}, our innovation lies in that we exploit the inductive biases to improve the state representation learned by the backbone of the agent's policy network, thus strengthening the ZSG capability of the agent's policy. Here, the inductive biases are twofold. First, the AOD problem should focus on the features concerning the target appearance and its spatial relation with surroundings, while information like color and texture is less important and should be filtered out. Next, because of the partial observability, we argue that the state representation should be extracted from the observation sequence instead of the single-view observation.

Since the proposed method is an improved work based on the Multistep Action Prediction (MAP) method \cite{17}, we call it Inductive Biases Enhanced Multistep Action Prediction (IBE-MAP), and its forward inference flowchart is demonstrated by Fig. \ref{fig7}. The upper part illustrates the whole AOD process. By making decisions within the time allowed, the UAV platform constantly optimizes its observing angle and derives the final ideal observation for target recognition.

During each decision step, the UAV platform moves to the new position according to the former action $\bm{a}_t$ and observes the region of interest. Next, we filter out irrelevant information from the observation by scene pre-decomposition operation. To achieve this, we use SAM \cite{29}, a powerful segmentation tool that can generate high-quality object masks given prompts like points or bounding boxes. Specifically, we first use part of the training data to fine-tune the original SAM and apply it to target segmentation given the observation $\bm{I}$ and the prompt of the bounding box. By taking a circle of sample points outside the target's contour as the prompts, we can also obtain the masks of the terrestrial objects around the target. In the simplified observations, as shown in Fig. \ref{fig7}, the pixels belonging to the target remain unchanged while the other objects are covered with their masks, whose gray values are randomly chosen between 0 and 1. In the next stage, the simplified images and bounding boxes are transformed into feature vectors and concatenated to form $\bm{g}_t$. The encoder used here is the backbone of a ResNet50 network pretrained on the ImageNet dataset. Later, the GRU network is utilized to learn an informative state representation $\bm{s}_t$ from the observation history. In the last step of decision making, for simplicity and fairness, we follow the practice in the literature \cite{17}, which proposes a novel DualingDQN algorithm, the network receives the state $\bm{s}_t$ and has two separate output channels,  respectively predicting action type and action range. By combining the values of these two dimensions, we could choose the optimal action $\bm{a}^*$ as the output action $\bm{a}_{t+1}$.

\section{EXPERIMENTS}

\subsection{Experimental Settings}
First, the UEVAVD dataset is divided into three parts, as shown in Table. \ref{tab1}. The training set is used to train the agent's policy network, and the hard test set is used for policy testing. Next, we choose an ImageNet-pretrained ResNet18 network as the classifier and finetune it with the unoccluded multi-view images of five targets. The comparison baselines are MAP and Memo-MAP, and the latter is the modified version of MAP, which incorporates the memory module for extracting state representation from the observation sequence while not including the scene pre-decomposition stage compared to IBE-MAP. The comparison results among them would verify the effectiveness of our innovations.

Since we aim to teach the agent how to actively search for the best view for target recognition with the shortest path, we adopt the evaluation metrics of return, recognition accuracy, and the overall path length within an episode in the following experiments. We mainly focus on the test-time performance and the generalization gap $GenGap(\pi)$ especially. The overall path length is decided by the number of decision steps and the path length per decision step.

By default, we assume the maximum decision steps $T_{max}$ is 5. The maximum action range $A$ is set to 4, and the discount factor $\gamma$ in return calculation is set to 0.9. The dimensions of $\bm{g}_t$, $\bm{h}_t$ and $\bm{s}_t$ are 544, 544, 512, respectively. In the reward function, $\xi_1=0.5$, $\xi_2=0.1$. Let the coefficient $\sigma$'s upperbound $C$ be 0.02, which can also be adjusted to balance movement cost and return value. During the training stage, we utilize the Adam optimizer with a learning rate 0.0001, and the total number of training episodes is 300000.

\subsection{Results and Analysis}
\begin{figure}[!t]
	\centering
	\includegraphics[width=3.4in]{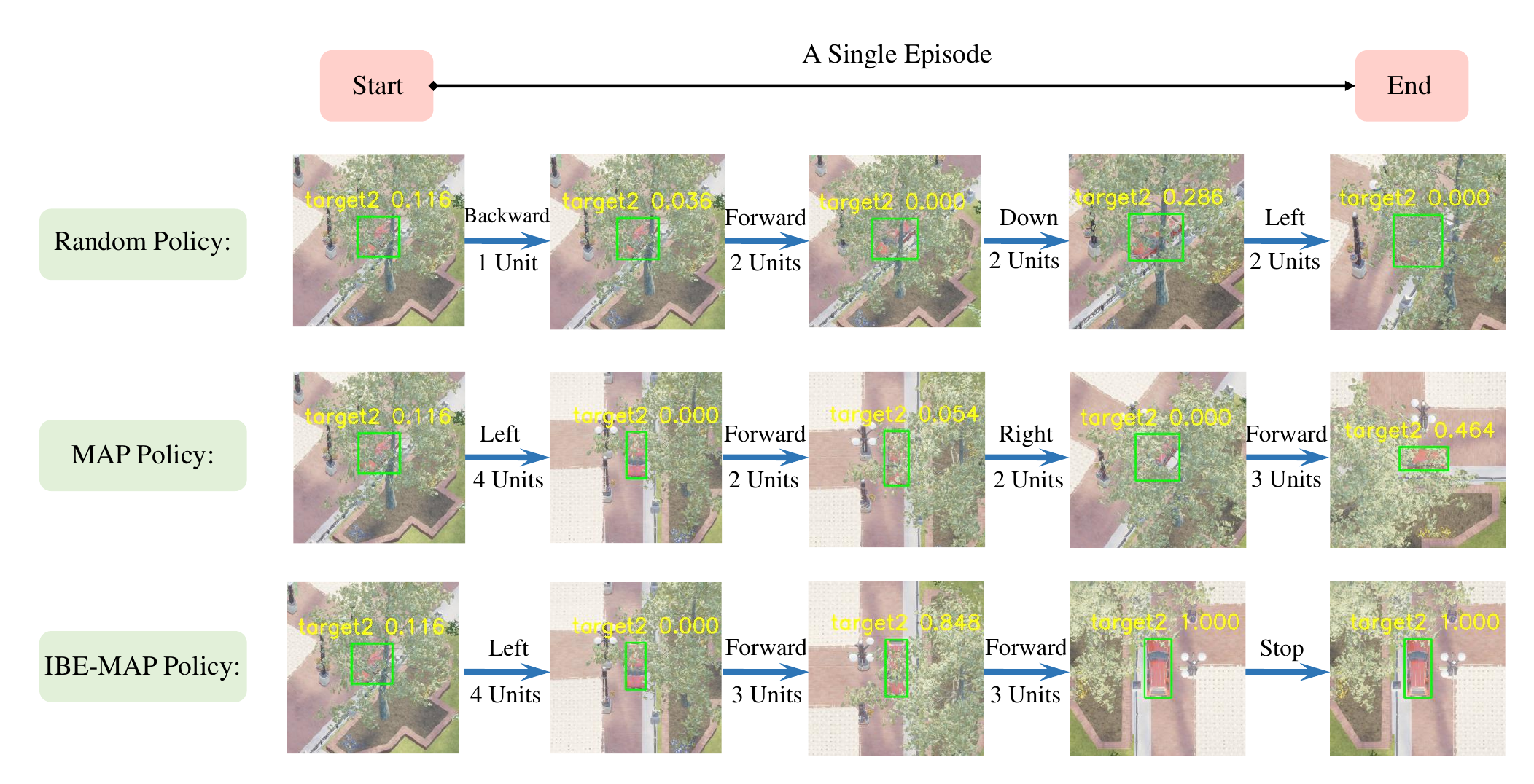}
	\caption{The observation sequences of different policies when tested on the hard test set.}
	\label{fig8}
\end{figure}

\textbf{Evaluation on the hard test set.} At first, with the training set of the UEVAVD dataset, we train the agent's policy using different AOD methods and test them with the hard test set. A straightforward comparison result is given in Fig \ref{fig8}. Given the same initial observation in the testing environment, three kinds of policy act differently within a single episode. The random policy arbitrarily gives action instructions, thus making the AOD task fail. The MAP policy succeeds in helping the UAV platform avoid the obstacle and get the correct recognition result. However, it clearly deviates from the optimal path. By contrast, the IBE-MAP policy guides the UAV to a better view at a lower movement cost. Once the agent finds a view that is sufficient for target recognition, it makes the early stop decision to increase efficiency in executing the AOD task.

\begin{figure}[!t]
	\centering
	\includegraphics[width=3.4in]{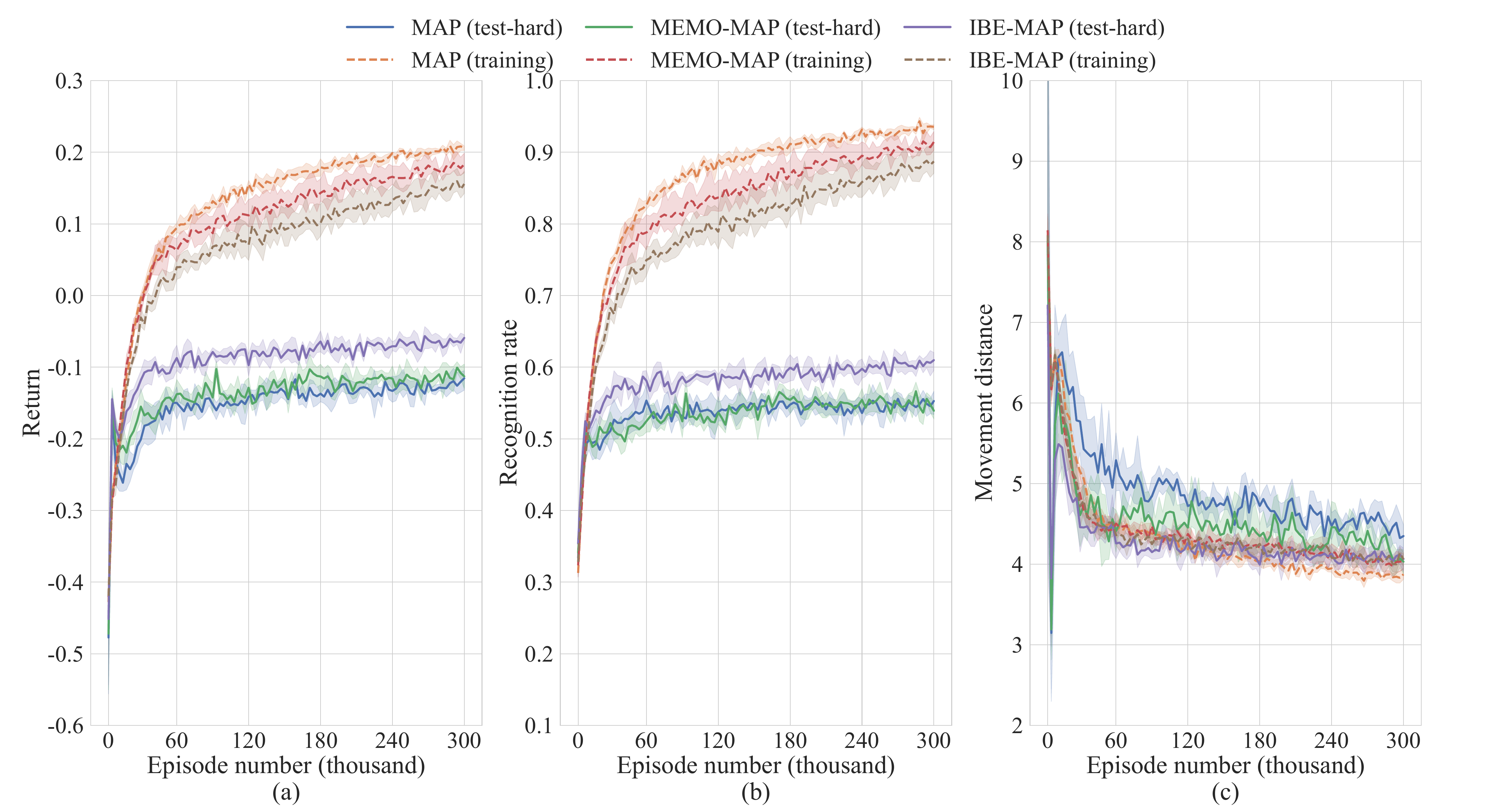}
	\caption{The evaluation results of three AOD methods concerning (a) return ,(b) recognition rate, (c) movement distance under the training and hard testing contexts.}
	\label{fig9}
\end{figure}

Then, we give the overall comparison result among three AOD methods in Fig. \ref{fig9}. It shows how their performances evolve regarding return, recognition rate, and movement distance during the training and test stage. To derive these colored lines, we first run the program under five random seeds to get five groups of fluctuating curves, and we smooth them using the sliding window averaging. Then, the tool named Seaborn is used for visualization, where the solid lines in the shaded area denote the mean values of those five smoothed curves, and the shaded area indicates the fluctuation range of the data, which is expressed by the confidence interval with a confidence level of 0.95. At the end of each training epoch, the test results are simultaneously calculated by evaluating the intermediate policies in the test environments. With the increase in training episodes, all the return curves under different policies go up, reflecting that they could help the agent get a better recognition result with a lower movement cost compared to the passive perception way.

Meanwhile, we can see that the policy derived by the IBE-MAP method holds the strongest generalization capability because, after convergence, its return value overwhelms the other two at the test time, and its $GenGap$ is the smallest. Although the original MAP performs better during training, its $GenGap$ is the largest, reflecting the overfitting problem. By incorporating the historical observations into the state representation, the policy derived by Memo-MAP is slightly better than that of MAP, whose movement distance is shorter and the $GenGap$ is smaller. The results above have validated the effectiveness of the innovations of scene pre-decomposition and memory based state estimation.

\textbf{Hyperparameter analysis.} In this section, we are going to analyze two Hyperparameters, $thre$ and $C$ in the reward function, to see their impacts on policy performance. All the evaluation results in this section are derived by testing the well-train policies on the hard test set, and the best results are shown in bold. Following the former works like \cite{11, 17}, $thre$ means the threshold for judging whether the classifier correctly recognizes the target, i.e., when the classifier gets a score higher than $thre$, it is seen as a successful classification. Obviously, $thre$ could affect the reward function, thus influencing the policy learned. Table. \ref{tab2} gives the policy evaluation results under different thresholds. We can see that as $thre$ grows, the return declines, the path length is prolonged, and its impact on the recognition accuracy is minor. The reason is that with a larger threshold, the agent is forced to move more frequently to find the `perfect' view, leading to a longer path length and a lower return. Since there are five kinds of targets, the minimum $thre$ is 0.2, and we find that the policy under this configuration performs best.
\begin{table}
	\caption{The agent's policy performance versus classification threshold}
	\label{tab2}
	\centering
	\renewcommand\arraystretch{1.25}
	\setlength{\tabcolsep}{0.8mm}{
		\begin{tabular}{|c|c|c|c|c|}
			\hline
			& $thre$=0.2 & $thre$=0.4 & $thre$=0.6 & $thre$=0.8 \\
			\hline
			Return & \textbf{-0.059±0.006} & -0.072±0.021 & -0.102±0.011 & -0.178±0.011 \\
			\hline
			Accuracy & \textbf{0.610±0.015} & 0.602±0.022 & 0.605±0.009 & 0.603±0.015 \\
			\hline
			Path Length & \textbf{4.065±0.169} & 4.160±0.171 & 4.211±0.152 & 4.505±0.190 \\
			\hline
		\end{tabular}
	}
\end{table}

In the reward function, $\sigma$ is a coefficient used to control the impact of action range on the reward value, thus balancing the movement cost and the profit of accuracy gain. By adjusting its upperbound $C$, we could see the variation trend of the policy's testing performances in Table. \ref{tab3}. When $C=0$, we only care whether the newly acquired imaging result can be recognized and pay no attention to how far the UAV moves to get that result. In other words, no further constraint is posed on the action range per step, so the total path length is the largest at this time when the agent can explore the environment more freely and gain better recognition accuracy. With the increasing of $C$, the constraint on action amplitude is heavier, shortening the path length while lowering the recognition rate. Clearly, there is a trade-off between accuracy and path length, so the setting of $C$ depends on their respective importance while executing the AOD task. 
\begin{table}
	\caption{The agent's policy performance versus hyperparameter $C$}
	\label{tab3}
	\centering
	\renewcommand\arraystretch{1.25}
	\setlength{\tabcolsep}{0.8mm}{
		\begin{tabular}{|c|c|c|c|c|}
			\hline
			& $C$=0 & $C$=0.02 & $C$=0.04 & $C$=0.06 \\
			\hline
			Return & \textbf{0.000±0.004} & -0.059±0.006 & -0.119±0.010 & -0.146±0.019 \\
			\hline
			Accuracy & \textbf{0.616±0.007} & 0.610±0.015 & 0.577±0.005 & 0.575±0.025 \\
			\hline
			Path Length & 4.717±0.160 & 4.065±0.169 & 3.509±0.104 & \textbf{3.147±0.076} \\
			\hline
		\end{tabular}
	}	
\end{table}

\section{CONCLUSIONS}

This paper releases a new dataset, UEVAVD, consisting of multi-view aerial images of vehicle targets under different terrains and occlusion conditions. Using the combination of these observations, one could simulate the process of the UAV's continuous observation while moving along the trajectory. We hope this dataset will be conducive to developing research on the AOD problem from the UAV's perspective. Besides, we proposed the IBE-MAP method, which incorporates two aspects of prior knowledge into the original MAP work, enabling the policy network to learn a better state representation. On this basis, the agent's policy can be generalized better to the testing environment. In future work, we will seek a more appropriate way to better the state representation for the policy network, thus further raising the test-time return value and narrowing the generalization gap.

\section*{ACKNOWLEDGMENT}

This work was not supported by the National Key Research and Development Program of China No. 2021YFB3100800, the National Natural Science Foundation of China under Grant 62376283, 61921001, 62022091, and 62201588, the Science and Technology Innovation Program of Hunan Province under Grant 2021RC3079.

%
%

\bibliographystyle{ieeetr}
\bibliography{reference}

\end{CJK}
\end{document}